\documentclass[conference]{IEEEtran}
\IEEEoverridecommandlockouts
\usepackage{amsmath,amssymb,amsfonts}
\usepackage{algorithmic}
\usepackage{graphicx}
\usepackage{biblatex}
\usepackage{textcomp}
\usepackage{xcolor}
\usepackage{subfigure}
\usepackage{xspace}
\usepackage{booktabs}
\usepackage{array}
\newcommand{\AGAMOTTO}{\textsc{AGAMOTTO}\xspace}

\newcommand{\added}[1]{\textcolor{black}{#1}}

\def\BibTeX{{\rm B\kern-.05em{\sc i\kern-.025em b}\kern-.08em
    T\kern-.1667em\lower.7ex\hbox{E}\kern-.125emX}}
\begin{document}

\title{AGAMOTTO: An AI-Enabled Framework for Time-Critical Tactical Course-of-Action Evaluation and Generation}

\author{\IEEEauthorblockN{Matthew L. Corbett}
\IEEEauthorblockA{\textit{U.S. Army Cyber Institute} \\
\textit{United States Military Academy}\\
West Point, New York \\
matthew.corbett@westpoint.edu}
}

\maketitle

\begin{abstract}
Current command and control (C2) platforms lack integrated simulation capabilities that would allow commanders to dynamically explore courses of action (COAs) during live operations. This paper presents ongoing research and contributions to the development of AGAMOTTO (Advanced Guidance and Analytics for Military Operations, Tactical Threats, and Outcomes) as part of the overarching NGC2 program, which advances the Army's efforts to modernize C2 systems, such as the Tactical Assault Kit (TAK). The paper outlines the current design, preliminary findings, and the remaining research tasks, with an emphasis on how the completed work will support tactical decision-making in future time-constrained environments. While the full system is still under development, initial development and results demonstrate the feasibility of AGAMOTTO. Early experiments highlight both the potential benefits and practical constraints of deploying AI-enabled tools in operational settings, and introduce lessons learned and a viable training strategy to make AI-generated COAs feasible in tactical scenarios.
\end{abstract}


\section{Introduction}
Modern military operations require rapid, data-driven decision-making under conditions of uncertainty. These challenges are likely to become more acute as both the enemy and friendly decision processes become driven by artificial intelligence tools\cite{adler_modernizing_nodate}, and decision timelines compress. At the tactical level, with limited resources and manpower, the ability to adapt to shortened decision timelines is hindered without access to AI-enabled decision support systems (DSSs). This creates an environment in which tactical decision-makers are at a disadvantage and require AI-integrated tools to overcome this disadvantage.

One potential avenue to integrate AI-enabled DSS into tactical operations is during planning. Tactical leaders must receive operational orders and synthesize them into a coherent, actionable plan. In the U.S. Army, at the company level and below, this process is called the Troop Leading Procedures (TLP) and requires a tactical leader to produce an order to their subordinate troops, including a plan to accomplish their mission and objectives. Their mission includes tactical tasks, specific requirements that outline the exact outcome desired with respect to friendly troops, the enemy, and the given terrain. This process can be labor-intensive, with little assistance to the tactical leader outside their own training and that of their subordinates. 

Integrating AI into this planning process can provide tactical leaders with advantages by reducing planning time and enabling task-focused optimization of outcomes. A directly accessible simulation environment could enable tactical leaders to digitally execute their plan without risking lives or other resources and to gain feedback before the start of the mission. Leveraging AI methods such as reinforcement learning (RL), this simulation environment could also serve as a platform for AI agents to explore potential solutions in search of an optimal outcome and recommend the corresponding path to the leader, thereby reducing cognitive and time requirements. However, no tools exist that can dynamically model the requirements of the tactical mission, including the specific terrain and enemy, focusing on the assigned tactical task. Such a tool would also need to be accessible to tactical leaders (i.e., integrated into their existing tactical systems) and generate recommended solutions within the unit's planning timelines, sometimes measured in only a few hours. 

Partly to address gaps such as this, the then Army Futures Command (AFC) commissioned the Next Generation C2 platform (NGC2)~\cite{army_futures_command_memorandum_2024}. Of the nine key characteristics outlined in the document, NGC2 ``must also include an organic simulation tool \dots.'' This paper outlines ongoing research and contributions to a tool to address these limitations as part of the overarching NGC2 program, which supports the Army's efforts to modernize C2 systems. This tool, named the Advanced Guidance and Analytics for Military Operations, Tactical Threats, and Outcomes (\AGAMOTTO) system, is designed to create a real-time feedback loop between an arbitrary C2 system (i.e., the NGC2 platform) and existing Army-funded simulation environments (i.e., One Semi-Automated Forces (OneSAF)) for: 1) simulation of planned Army operations to aid in rehearsals and planning, and 2) predictive analysis of current tactical situations by simulating and visualizing potential enemy actions based on known or templated enemy situation. 

In this paper, we outline the \AGAMOTTO system, and discuss the limits of existing RL techniques applied to similar problem sets. Additionally, we outline other state-of-the-art RL methods that may be more generally applicable to the mission and time requirements in tactical operations. Specifically, this paper introduces:

\begin{itemize}
    \item A novel methodology for integrating simulation and AI into currently fielded tactical C2 systems to improve tactical outcomes and planning.
    \item Initial evaluation results for a task-focused RL reward system, training AI agents to optimize tactical outcomes according to specific mission requirements. 
    \item A discussion of the current technical difficulties of such systems, and initial work to overcome them.
    \item An outline for a way forward to fully develop and integrate such a system into tactical C2 systems, such as NGC2

\end{itemize}

\section{Related Works and Key Insights}

\subsection{AI-Integrated Decision Support Systems}
Due in no small part to White House guidance, the Department of War (DoW) has accelerated programs to integrate AI in military DSS~\cite{us_department_of_war_artificial_nodate, unied_states_executive_branch_ai_nodate}. Projects such as Thunderforge support strategic, theater-level planning through large language models and battle management systems, such as Anduril's Lattice~\cite{defense_innovation_unit_thunderforge_nodate}. Other projects have been linked to strategic and theater targeting, as well as battlespace management~\cite{caballero_teston_role_2026,shaikh_iran_nodate}. These systems are designed to assist in large-scale planning efforts led by senior commanders and large staffs. The U.S. Army War College has experimented with AI for wargaming, a method for evaluating the efficacy of plans before execution, but this is generally not done at the tactical level~\cite{timothy_j_williams_ai-enabled_2026}. There has been very little research on integrating simulation and optimization capabilities at the tactical level, creating a gap in AI integration efforts across the operational force. 

\subsection{Reinforcement Learning for Tactical Optimization}
RL has been identified as a potential avenue to address the gap in AI-integrated tactical DSS~\cite{adler_modernizing_nodate}. RL has achieved breakthroughs in complex decision-making in games, achieving superhuman outcomes~\cite{silver_mastering_2016, vinyals_grandmaster_2019, openai_dota_2019, mnih_playing_2013}. 

RL is an AI learning framework in which an agent learns to select actions in an environment in order to maximize cumulative reward. An RL problem is typically formalized as a Markov Decision Process (MDP)
\(\mathcal{M} = (\mathcal{S}, \mathcal{A}, P, R, \gamma)\), where:
\begin{itemize}
    \item \(\mathcal{S}\) is the set of states,
    \item \(\mathcal{A}\) is the set of actions,
    \item \(P(s' \mid s, a)\) is the transition probability,
    \item \(R(s,a)\) is the expected immediate reward,
    \item \(\gamma \in [0,1)\) is the discount factor.
\end{itemize}

At each time step \(t\), the agent observes a state \(s_t \in \mathcal{S}\),
selects an action \(a_t \in \mathcal{A}\) according to a policy
\(\pi(a \mid s)\), receives a reward \(r_t\), and transitions to a new state
\(s_{t+1}\) according to \(P\).

The objective of RL is to find an optimal policy \(\pi^*\) that maximizes the expected discounted return
\begin{align}
G_t = \sum_{k=0}^{\infty} \gamma^{k} r_{t+k}.
\end{align}

In our use case, a state could be a set of friendly and enemy positions. Actions could include movement (i.e., a single ``step'' of a fraction of a meter in a given direction), firing a weapon, or other tactical actions. An optimal policy would allow an agent to traverse a digital environment in such a way to maximize their rewards, defined arbitrarily but likely tied to control of terrain, destruction of enemies, and loss of friendly troops. 

Recent work has sought to leverage this technology for military applications, using high-fidelity military simulations as training environments~\cite {dimitriu_reinforcement_2024, black_mastering_2024}.  Goecks et al. have even attempted to integrate mission tasks into the reward system for RL training, demonstrating the feasibility and advantages of this approach~\cite{goecks_games_2023}. However, most work, including work by Goecks et al, uses pixel data from the simulation itself for training. This is infeasible in a live tactical environment, such as our use case, because it can only train on what is shown on the simulation screen rather than on a higher-fidelity method, such as a combat simulator. This distance between the real world and the simulation highlights the so-called ``sim-to-real'' gap faced in modern military simulations~\cite{nack_navigating_nodate}.

\subsection{Key Insights}
Military simulators, such as the Rapid Integration and Development Environment (RIDE), can provide a simulation and training platform that uses realistic combat physics and terrain~\cite{hartholt_introducing_nodate}. This can overcome the lack of high-fidelity training environments for RL agents and address the gap in AI-enabled tactical DSS. When combined with RL packages, these simulation environments can replicate the tactical leader's real-world conditions and enable RL to explore realistic terrain, enemy engagements, and friendly actions. When combined with information from existing tactical DSS, such as the Tactical Assault Kit (TAK), which is widely fielded today, this training backbone can optimize outcomes against the enemy envisioned in the tactical leader's order, on the actual terrain, and informed by the leader's tactical objectives and tasks. 

Using these insights, we now seek to outline a system for AI-integrated tactical DSS, which we call \AGAMOTTO. 

\section{Technical Requirements}

\begin{figure}
    \centering
    \includegraphics[width=0.85\linewidth]{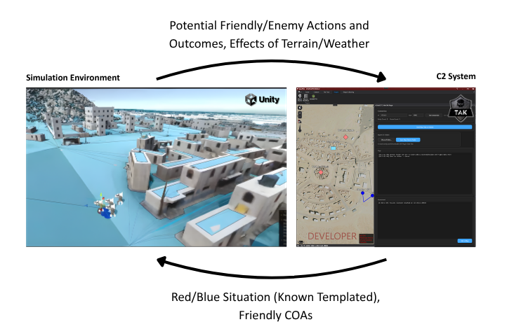}
    \caption{An abstract representation of the \AGAMOTTO design. With further research and development, \AGAMOTTO can serve as a simulation and COA-development tool for tactical leaders.}
    \label{fig:abstract}
\end{figure}

\AGAMOTTO is envisioned as a prototype, proof-of-concept system that provides tactical leaders with access to a simulation and COA generation tool as part of C2 system functionality, as envisioned in NGC2, to aid in the planning, evaluation, and execution of tactical operations. As NGC2 is still under active development, \AGAMOTTO must be modular, providing a platform-agnostic method to encapsulate and transfer data between arbitrary C2 systems and simulation environments. Figure~\ref{fig:abstract} gives a high-level overview of this design. \AGAMOTTO must retrieve current operational and planning data, such as terrain, mission, and enemy locations, and package this in a standardized format. The simulation and training environment must then either ``execute'' the mission as planned or simulate a range of potential enemy and friendly actions to compare and optimize COAs. \AGAMOTTO must then return the simulation results or optimized COA for further integration into the tactical leaders' plan. 

 In simulation mode, \AGAMOTTO must recreate planned missions, intents, checkpoints, and movement routes to provide the user ``playback'' of their planned mission, including simulated effects of enemy action and expected weather. This will aid in rehearsal, wargaming, and mission execution. In COA generation mode, \AGAMOTTO will ingest current known friendly locations, weather, terrain, unit strengths, and expected missions, then pair these with expected enemy units and locations, and simulate multiple potential enemy and friendly outcomes. This will aid the tactical leader in generating, evaluating, and planning tactical courses of action in real time. These two use cases fundamentally differ in their understanding of the environment, certainty, and visualization. Uncertain outcomes, such as in prediction mode, should be visualized as a range of possible outcomes conditioned on factors such as enemy routes, intent, and strength. In contrast, the wargaming mode assumes relative certainty from a planned friendly operation. It provides the user with a method to visualize their plan, including the effects of terrain and enemy actions. 

\subsection{Challenges}
The creation of a system such as \AGAMOTTO requires developing several new capabilities. Developing a platform-agnostic method for transferring tactical data will require synthesizing, serializing, and transferring friendly and enemy locations, expected actions, and weather across two disparate C2 and simulation platforms. This will necessitate creating a ``wrapper'' to retrieve the serialized information from the C2 system, create a synthetic scenario to match it, run the simulation, and return the results to the C2 system for visualization. Additionally, the Army's C2 systems lack an effective method for visualizing uncertainty in tactical environments. This is required to convey the uncertainty in the range of potential outcomes generated by \AGAMOTTO. 

Finally, but not least pressing, \AGAMOTTO will require custom-trained agents tailored to arbitrarily defined terrain, enemy, and mission objectives. Unlike previous work, the rules of our ``game'' will be scenario-specific. RL training in complex environments is challenging under real-time requirements~\cite{black_mastering_2024}. RL training can take many hours, days, or weeks, as shown in early work optimizing game outcomes~\cite{mnih_playing_2013,openai_dota_2019}. This is not feasible in tactical environments, where planning is often time-constrained to hours. Optimizations are required to reduce these training times and make \AGAMOTTO viable. A viable prototype system must overcome these challenges to prove that real-time AI-integrated DSS is viable at the tactical level.

\section{\AGAMOTTO Implementation}

\AGAMOTTO is intended to be a modular system to enable deployment across a variety of NGC2 solutions that could be fielded in the near future. For the tactical planning tool that will serve as the front end for the tactical leader, TAK is a widely fielded visualization, targeting, and planning system in use by the joint force today~\cite{noauthor_takgov_nodate}. TAK also exposes an SDK on some variants of its software, and is a government-off-the-shelf (GOTS) system. TAK mission data (i.e., the mission planning data we seek for our experimentation) can be retrieved and distributed as ``cursor on target'' messages, a standardized military message format~\cite{butler_developers_2005}. Using such a format in our implementation means that, as long as the front-end and simulation backend can use this standardized format, the front-end and backend systems will remain platform-agnostic.   

RIDE was chosen as the simulation backbone because it is a software package and integrates with Unity RL packages such as MLAgents~\cite{hartholt_introducing_nodate, unity_corporation_what_nodate}. This provides both simulation functionality and the ability to harness the simulator's physics and combat logic for RL agent training.  RIDE also provides terrain representations of many areas, including major training centers such as the National Training Center at Fort Irwin, California. This terrain not only has real-world topology, but also a 3D representation of the buildings and other infrastructure. Note that RIDE supports simulation of weather effects, but we do not use that functionality in the initial prototype~\cite{usc_ground_nodate}. 

To ensure that our prototype can operate on separate networks, a middleware solution is required. Placing a middleware system on a simulation host allows requests and response messages from \AGAMOTTO to be sent, provided there is a viable path between the frontend and backend systems, subject to applicable network permissions, firewall rules, and other configuration details. 

We chose to begin with a small subset of the many possible missions, terrain, and enemy variations, and to pick a single geographic location for two potential mission tasks. This was done to provide an initial assessment of feasibility and efficacy before scaling to a larger prototype. The location for agent training is the NTC training area, specifically a mockup of a large rural village fictitiously named Razish. A squad-sized unit comprising both friendly and enemy troops, each with about 10 soldiers, is used in our scenario. These soldiers are equipped with small arms and weapons appropriate to their size and threat alignment. No vehicles, artillery, or aircraft are simulated, although RIDE can simulate these capabilities at a small scale in future work. 

\begin{figure}
\centering     
\subfigure[A screenshot from the AGAMOTTO frontend, implemented on TAK. The training area in the image is the Razish village at the NTC at Ft. Irwin, CA. ]{\label{fig:a}\includegraphics[width=0.85\linewidth]{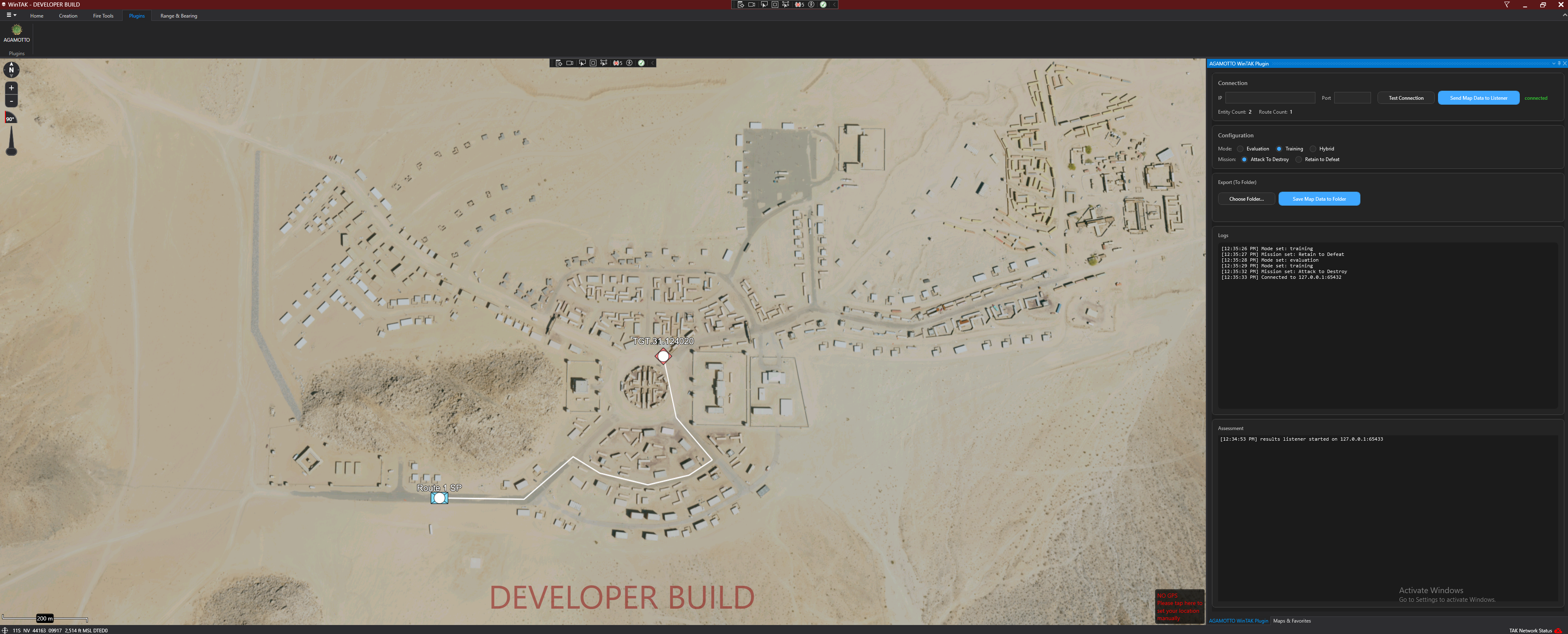}}
\subfigure[A screenshot from the AGAMOTTO backend, implemented with RIDE and Unity. The training area in the image is the same village as in \ref{fig:a}. The screenshot was taken as the friendly squad (center of the screenshot) progressed along its directed route (the white route line in \ref{fig:a}) toward its objective (an enemy squad out of view, but depicted as a red diamond on \ref{fig:a}).]{\label{fig:b}\includegraphics[width=0.85\linewidth]{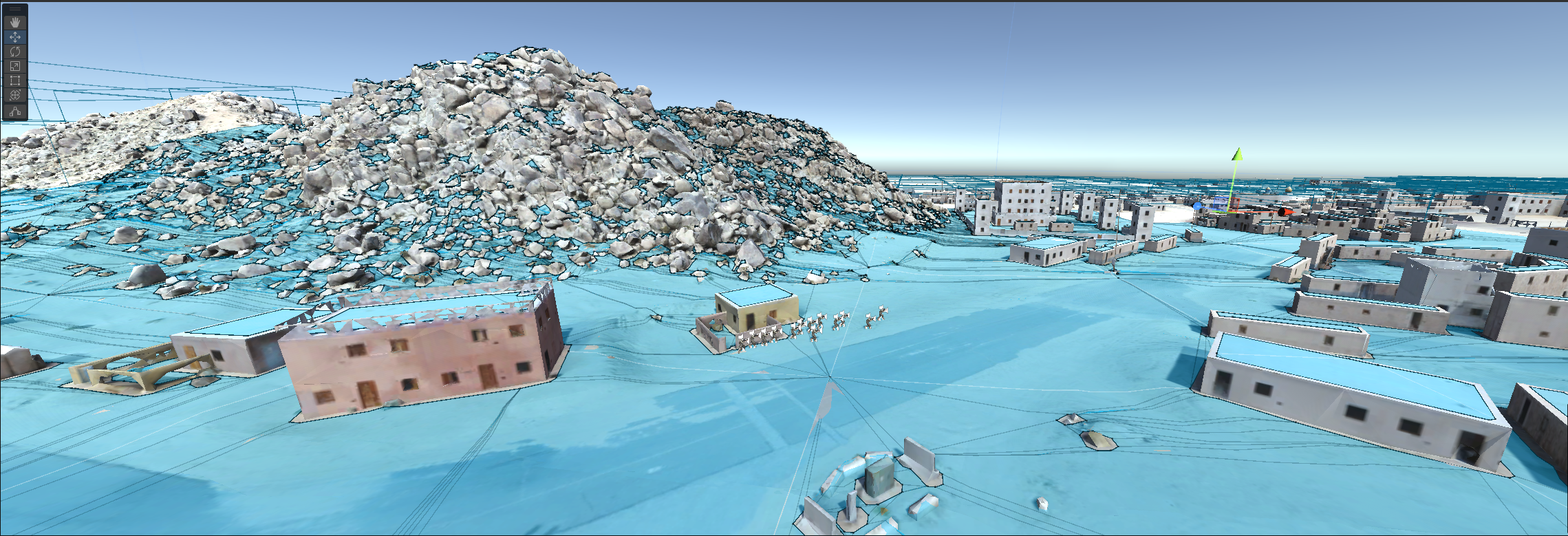}}
\caption{A side-by-side view of the functionality of AGAMOTTO. \ref{fig:a} and \ref{fig:b} show the same geographic area, with \ref{fig:a} showing the tactical leader's view and \ref{fig:b} showing the simulation backend.}
\label{fig:AGAMOTTOSidebySide}
\end{figure}

Initial RL training is performed using the Soft Actor-Critic (SAC) deep reinforcement learning (DRL) method for its general stability, sample efficiency, and built-in exploration incentive~\cite {haarnoja_soft_2018, williams_simple_1992}. These qualities are expected to assist with efficient training in the very large state space of our life-sized training environment, encouraging model updates for new and unexplored paths, which translates directly to paths in the digital terrain of RIDE. Razish and the NTC are both very large environments in the real and digital worlds, at roughly 1,200 square miles (about the size of Rhode Island). The RIDE MLAgents package provides pre-built SAC functionality, with support for customizing step counts and a variety of other training hyperparameters. Agents are provided with a simple observation space, namely the 3D Cartesian coordinates of their current location and the expected objective location. This objective can be a location or an enemy force, depending on the mission. Finally, to improve training speed, we leverage RIDE's accelerated scenario timescale options to prevent training from proceeding at the speed of actual ground combat, theoretically making the learning process feasible for tactical use. Figure \ref{fig:AGAMOTTOSidebySide} illustrates both the TAK frontend and the RIDE/Unity backend.

Finally, the mission-driven reward structure for our agents is a core contribution of this work. Initial attempts to design these values focused on one tactical task and associated purpose: attack to destroy (ATD)~\cite{us_army_army_2026}. An attack to destroy is an offensive operation that requires the tactical leader to render an enemy force ineffective, typically measured by reducing enemy numbers. This is a simple, straightforward task and purpose that is easily measured to calculate rewards. 

Positive rewards are given for eliminating enemies in ATD, whereas negative rewards are given for losing friendly troops. Additionally, smaller negative rewards are assigned at each step of the mission's execution, incentivizing faster solutions. We provide more detail on these rewards in Table~\ref{tab:unity-extrinsic}.

\begin{table}[t]
\centering
\caption{\added{Unity extrinsic environmental rewards for Attack-to-Destroy.}}
\label{tab:unity-extrinsic}
\small
\begin{tabular}{@{}lrl@{}}
\toprule
\textbf{Event} & \textbf{Reward} & \textbf{Notes} \\
\midrule
Enemy unit killed & $+5$ & Episode success \\
Friendly squad killed & $-5$ & Episode failure \\
Single enemy killed & $+0.45$ & Per death \\
Single friendly killed & $-0.50$ & Per death\\
Closer to enemy & $+0.002$ & Per decision \\
Farther from enemy & $-0.0008$ & Per decision \\
Obstacle collision & $-0.002$ & Per collision \\
Max steps ($N_{\mathrm{ep}}{=}5000$) & $0$ & Episode end \\
\bottomrule
\end{tabular}
\end{table}

\section{Early Results}
\label{sec:results}
Training was conducted on an NVIDIA RTX 5090 GPU and a Dell Alienware Area-51. This is an off-the-shelf system priced at around \$4000 to \$5000, depending on specific options. We use this system, rather than a more capable compute cluster, to better simulate a locally deployed solution at the tactical level. While cloud compute solutions may be available to the unit, we believe a local GPU with a capable host is a reasonable minimum baseline.

\subsection{\added{Evaluation of AGAMOTTO COA Generation}}
\label{sec:evaluation}

\added{We evaluate AGAMOTTO along three axes aligned with NGC2 planning needs: (i)~\emph{C2--simulation fidelity}---whether tactical data encoded as CoT messages reproduce the intended scenario in RIDE; (ii)~\emph{learning efficiency}---whether mission-focused RL training reaches usable policies within tactically relevant wall-clock and step budgets on local hardware; and (iii)~\emph{mission outcome quality}---whether learned behavior improves ATD outcomes relative to ablated training variants. A fourth axis---cognitive and time burden for tactical leaders---is reserved for a planned in-person study.}

\subsection{\added{Evaluation Conditions}}
\added{While each tactical scenario is different, we must assume a maximum total time allowed to tactical leaders in order to evaluate the feasibility of AGAMOTTO's design. Additionally, our RIDE does not claim to emulate the movements of live units exactly, defaulting to pre-defined tactical formations (e.g., wedge or file). This oversimplification removes the nuance of scenario-based movement like bounding or covering, but allows for the simulation to be tractable. For our evaluation, we use the conditions below:  }
\added{
\begin{itemize}
  \item Simulation results must be returned in less than 6 hours to be viable. While every scenario is different, at the squad and platoon levels, this is generally the order of magnitude for planning timelines.
  \item We default to tactical movement in fire team wedges aligned in a squad file. There are certainly more nuanced approaches, but we believe this is a good generalization of common movement techniques.
\end{itemize}
}

\subsection{\added{Evaluation Objectives and Research Questions}}
\added{
\begin{itemize}
  \item \textbf{RQ1 (Integration):} Can AGAMOTTO ingest CoT-derived friendly and enemy dispositions (and routes when present) without manual re-entry into the simulator?
  \item \textbf{RQ2 (Training feasibility):} Under fixed hardware (RTX 5090, local deployment per Sec.~\ref{sec:results}), does SAC training exhibit stable improvement in cumulative return and mission metrics before the maximum step budget and tactical planning time assumptions (e.g., 6-hour maximum for COA generation results generation and feedback).
  \item \textbf{RQ3 (Reward design):} Do mission-task rewards (enemy elimination, friendly loss penalties, step penalties, etc.) measurably affect learned policies compared to ablated reward signals?
\end{itemize}
}

\subsection{AGAMOTTO in Simulation Mode (RQ1)}
Initial experiments with AGAMOTTO in simulation mode demonstrated that a single friendly squad can be replicated in RIDE and, when instructed to move along a predetermined avenue of approach (AoA) set by the tactical leader, can advance along it. This was not unexpected, as simulation mode relies solely on mirroring the expected mission and tactical situation in the simulation backend. Using TAK's route-planning functionality, we assigned a unit to a route, assuming the route's start is near the intended unit in the digital space.  However, it is noteworthy that certain tactical movement techniques are not replicated in RIDE, such as reconnaissance. For example, in a real-world tactical situation, even a unit as small as a squad may send a few soldiers to assess road or terrain conditions or confirm the expected location of the enemy. A real squad may also employ tools such as drones or indirect-fire systems when necessary. AGAMOTTO, in its current RIDE implementation, does not provide a method to simulate these intricate tools. However, simple missions can be simulated, or the tactical leader can use this simulation to estimate movement time requirements and to confirm expected engagement ranges using RIDE's built-in tactical engagement systems.

\subsection{AGAMOTTO in COA Generation Mode (RQ 2 and 3)}
In a potentially more compelling use case, we now discuss the results of AGAMOTTO's COA generation mode. This functionality was tested using the same tactical situation as in the simulation tests, with the friendly squad given an ATD task and purpose. After the creation of the applicable units, the friendly unit RL agent can explore paths and actions in accordance with its SAC learning procedures. 

It was initially clear that allowing RL agents to physically explore an area as large as NTC, starting with inherently random movement paths, was time-consuming. Early attempts at training convergence exceeded 8 hours and typically ended at the maximum step and episode counts rather than at convergence. \added{With SAC learning occurring in the Unity simulation environment at 50 times the normal speed, the total time to the maximum step count was 3.8 hours, as shown in the first line of Table~\ref{tab:ablation}. After nearly 4 hours, rewards had converged, but the friendly squad never met the conditions for victory by destroying the enemy unit. To further analyze this shortcoming, we conducted an ablation study of SAC learning variables.}

\subsubsection{\added{Ablation Testing}}
\added{Table~\ref{tab:reward-ablations} summarizes the 5 ablation tests, including the baseline SAC parameters. Figure~\ref{fig:reward-analysis} shows the episodic rewards over each ablation and episode. Ablation parameters were chosen to isolate individual SAC parameters and measure their impact on training time and rewards.}

\begin{table*}[t]
\centering
\caption{\added{Intrinsic and extrinsic reward signals by evaluation ablation.}}
\label{tab:reward-ablations}
\small
\setlength{\tabcolsep}{4pt}
\begin{tabular}{@{}l
  >{\raggedright\arraybackslash}p{2.35cm}
  >{\raggedright\arraybackslash}p{1.85cm}
  >{\raggedright\arraybackslash}p{4.85cm}
  >{\raggedright\arraybackslash}p{3.0cm}@{}}
\toprule
\textbf{Ablation} &
\textbf{Intrinsic Rewards} &
\textbf{Extrinsic Rewards (SAC)} &
\textbf{Unity Environment Rewards} &
\textbf{Isolated change} \\
\midrule
Baseline SAC &
Curiosity $\gamma{=}0.99$, strength $0.05$ &
$\gamma{=}0.99$, strength $1.0$ &
Full ATD (Tab.~\ref{tab:unity-extrinsic}) &
--- \\
No Curiosity &
Disabled &
$\gamma{=}0.99$, strength $1.0$ &
Full ATD &
No curiosity \\
Lower LR &
Same as baseline &
$\gamma{=}0.99$, Strength $1.0$ &
Full ATD &
SAC LR $10^{-4}$ \\
Shorter Horizon &
Same as baseline &
$\gamma{=}0.99$, Strength $1.0$ &
Full ATD &
SAC $H{=}256$ \\
No Squad Rewards &
Same as baseline &
$\gamma{=}0.99$, Strength $1.0$ &
Terminals + distance + collision only &
Squad kill/loss $0$ \\
\bottomrule
\end{tabular}
\end{table*}

\subsubsection{\added{Ablation Testing Results}}
\begin{figure}[!h]
\centering
\includegraphics[width=0.95\linewidth]{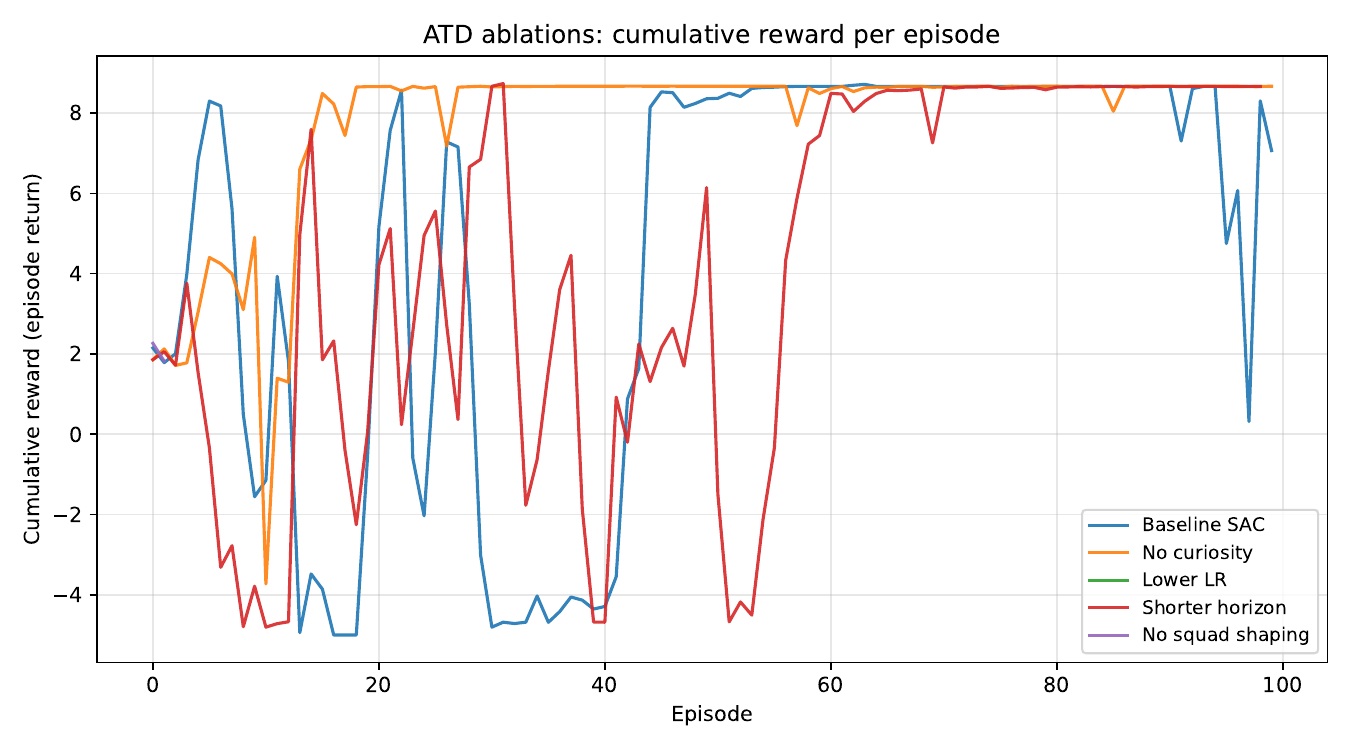}
\vspace{0.8em}
\caption{\added{AGAMOTTO cumulative rewards for the five ablations. }}
\label{fig:reward-analysis}
\end{figure}

\added{Table~\ref{tab:ablation} summarizes ablation outcomes. While the ablation identified methods to marginally increase training rewards, no ablation test succeeded in producing the average enemy-killed-to-friendly-survival ratio that would define success in a tactical ATD mission. Notably, the training times for each of these ablation was inside of our 6-hour evaluation requirements. The obvious convergence shown in Fig.~\ref{fig:reward-analysis} indicates that learning occurred, but it failed to find the optimal paths that led to optimal kill/death ratios. This is very likely due to an early convergence of locally optimal paths that do not include the globally optimal paths a tactical leader may create through manual planning. For more information on the effects of this, please see Section~\ref{sec:discussion}.}



\begin{table*}[!h]
\centering
\caption{\added{ATD training ablation summary}}
\label{tab:ablation}
\small
\begin{tabular}{@{}lcccc@{}}
\toprule
Condition & Average rewards (last 50\% of episodes) & Time (min) & Enemy kills/ep & Friendly losses/ep \\
\midrule
Baseline SAC & 7.75 & 228.9 & 0.00 & 0.00 \\
No curiosity & 8.20 & \textbf{155.2} & 0.00 & 0.00 \\
Lower LR & 1.77 & 209.1 & 0.00 & 1.00 \\
Shorter horizon & \textbf{8.66} & 211.6 & 0.00 & 0.00 \\
No squad shaping & 2.04 & 211.0 & \textbf{1.00} & 0.50 \\
\bottomrule
\end{tabular}
\end{table*}

\subsection{\AGAMOTTO in Hybrid Simulation/COA Generation Mode}

Given these results, we must explore new strategies to theoretically improve RL methods, such as SAC, for tactical COA generation. Generative Adversarial Imitation Learning (GAIL) leverages so-called ``expert trajectories'' to inform RL agents and streamline training. In practice, GAIL-trained agents can be provided with many examples of near-optimal behavior and then use these examples as a baseline. In our use case, it is a critical insight that we indeed do have a single ``expert trajectory'', developed by the tactical leader as part of the planning process. This trajectory is the initial plan, developed by the tactical leader to accomplish their assigned tactical objectives. It is intuitive that this example would be better than a truly random action set, like that of an untrained RL policy, but not optimal in terms of rewards and outcomes.

Given our expert trajectory \(\tau_E = \{(s_t, a_t)\}\), a discriminator
\(D_\theta(s,a)\) is trained to distinguish expert state--action pairs
from those generated by the policy \(\pi_\phi\). The policy is trained to
produce trajectories that the discriminator classifies as expert-like.

The GAIL objective is

\begin{align}
\min_{\pi_\phi} \max_{D_\theta} \;
&\mathbb{E}_{(s,a)\sim \pi_E}\!\left[\log D_\theta(s,a)\right] \\
&+ \mathbb{E}_{(s,a)\sim \pi_\phi}\!\left[\log\left(1 - D_\theta(s,a)\right)\right]
- \lambda \, \mathcal{H}(\pi_\phi)
\end{align}

where \(\mathcal{H}(\pi_\phi)\) is the causal entropy of the policy,
encouraging exploration and preventing premature convergence at some local optima. In our use case, this prevents focusing on a single advantageous COA when a different, more optimal COA may still exist. 

The discriminator provides a learned reward signal
\begin{align}
r_\theta(s,a) = -\log\left(1 - D_\theta(s,a)\right),
\end{align}
which is used by a reinforcement learning algorithm to update the policy
parameters \(\phi\). At convergence, the policy \(\pi_\phi\) matches the
expert's occupancy measure and reproduces expert-like behavior without
explicitly recovering the expert's reward function.

It is worth noting that GAIL does \emph{not} use the environment reward function (e.g., mission-focused rewards) in the same way as our SAC methods do. GAIL trains a policy by imitating the actions performed by the ``expert'' actor, in this case, those in our tactical leader's plan. GAIL does not generate actions based on a mission task and purpose, but rather imitates the leader's actions designed to achieve those goals. This is an indirect way to model an optimal outcome based on those rewards, so we seek a more direct and efficient path to COA generation.

In contrast, advantage-weighted actor-critic (AWAC) combines the expert example with the rewards of the pure-RL example (i.e., our mission-focused reward system)~\cite{nair_awac_2021}. AWAC assigns high weights to the expert example but does not simply imitate it. 

AWAC updates its policy by solving:
\begin{align}
\max_{\pi_\phi} 
\mathbb{E}_{(s,a)\sim \mathcal{D}}
\left[
\exp\!\left(\frac{A^{\pi}(s,a)}{\lambda}\right)
\log \pi_\phi(a\mid s)
\right]
\end{align}
where $A^{\pi}(s,a)=Q_\theta(s,a)-V_\theta(s)$ is the critic-based advantage estimate.

The critic is trained via temporal-difference (TD) learning:
\begin{align}
\min_{\theta}
\mathbb{E}
\left[
\left(
Q_\theta(s,a)
-
\left(
r + \gamma \, \mathbb{E}_{a'\sim \pi_\phi(\cdot\mid s')}
Q_{\theta^-}(s',a')
\right)
\right)^2
\right]    
\end{align}

This objective performs advantage-weighted behavioral cloning, allowing the policy to imitate high-advantage actions (i.e., our tactical leader's example) while still improving beyond the demonstrations. AWAC directly uses the mission-focused reward function in its optimization strategy, unlike GAIL. 

\added{While currently untested, methods such as AWAC or GAIL may be a way to find tactically optimal outcomes in feasible amounts of time. Issues with early convergence in pure SAC methods show the difficulty of allowing RL agents to learn in massive, highly complex environments such as the Razish training village at NTC. }

\section{Discussion}
\label{sec:discussion}
\AGAMOTTO is designed to provide tactical leaders with a tool to simulate the outcomes of their plans and a method to generate optimal outcomes given a templated mission and enemy. While the prototype is still under active development, we can report a few interim lessons learned. 

\subsection{Real Time Simulation is Feasible, but Benefits are Still Unknown}
\AGAMOTTO was initially successful with replicating a TAK-based scenario into Unity using RIDE. While a true in-person evaluation is forthcoming, the methodology outlined in this paper has been shown to be feasible, at least at squad scale, in the initial testing. However, RIDE lacks simulation fidelity that can alter outcomes, such as the ability of combat agents to enter or take cover in buildings or call for indirect fires, if these were planned as part of the operation. This can make the outputs shallow, but not necessarily useless. Existing functionality can depict movement times, engagement ranges, and potential casualty outcomes of a simulated operation. Even without the added fidelity of a more realistic simulation environment, these can assist a tactical leader with planning for operation lead times, casualty evacuation, and firing locations. 

\subsection{Pure RL is likely Insufficient for Real Time COA Generation for Tactical Leaders}

We initially attempted to generate novel COAs from a templated friendly and enemy location, given an assigned task and purpose. With a simple ATD task in a highly complex training environment, the agents converged on a locally optimal policy in less than the maximum required time, but likely failed to converge on the globally optimal policy. We believe these issues make SAC-based  RL techniques unlikely to be used in tactical settings. To that end, we have explored other RL-based methods that may benefit from the critical insight that we have an expert example in the tactical leader's planned mission. 

\subsection{Hybrid Imitation/RL Techniques May Provide a Feasible COA Generation Method}
AWAC creates optimized action policies using expert examples, such as the planned operation provided by the tactical leader in our use case. While development and evaluation are ongoing, we hypothesize that methods such as AWAC can reduce the time requirements for pure-RL COA generation and provide a viable real-time (or near-real-time) COA generation solution for tactical leaders. If this hypothesis proves valid, we have a solution that a tactical leader can evaluate for efficacy by examining changes in the cognitive and time requirements of planning. Further evaluations can also use the existing RIDE simulation environment to model casualty and mission competition rates, in preparation for a larger in-person evaluation. 

\section{Conclusion}

Initial development and evaluation of \AGAMOTTO show that real-time (or near-real-time) simulation of tactical outcomes is feasible. It also shows that integrating RL for COA generation is challenging and likely requires compromises and the development of new techniques. COA generation techniques that combine pure RL methods with imitation learning show promise for reducing the prohibitive time requirements of pure RL methods, such as SAC. Future work to integrate these methods and conduct a full simulated and in-person evaluation is ongoing.

\printbibliography

\end{document}